\documentclass[12pt, a4paper]{article}

\usepackage{microtype}
\usepackage{courier}
\usepackage{inconsolata}

\usepackage{xcolor}
\usepackage{xspace}

\usepackage[authoryear, sort&compress, round]{natbib}
\usepackage{graphicx}
\usepackage[small]{caption}

\usepackage[hyphens]{url}
\usepackage[hidelinks]{hyperref}
\hypersetup{breaklinks=true}
\usepackage{cite}

\usepackage{titlesec}
\usepackage{authblk}

\usepackage{changepage} 
\usepackage{parskip}    

\usepackage{appendix}

\newcommand{\promptfont}{\fontfamily{fi4}\selectfont}
\newcommand{\promptfontsize}{\fontsize{11pt}{14pt}\selectfont}
\definecolor{promptcolour}{RGB}{230,97,0}
\newenvironment{prompt}{
    \begin{adjustwidth}{1.5em}{1.5em}
    \promptfont\promptfontsize
    User:
    \color{promptcolour}
    \setlength{\parskip}{1ex}
    \noindent
}{
    \end{adjustwidth}
}

\newcommand{\chatfont}{\fontfamily{fi4}\selectfont}
\newcommand{\chatfontsize}{\fontsize{11pt}{14pt}\selectfont}
\definecolor{chatcolour}{RGB}{93,58,155}
\newenvironment{chat}{
    \begin{adjustwidth}{1.5em}{1.5em}
    \chatfont\chatfontsize
    ChatGPT:
    \color{chatcolour}
    \setlength{\parskip}{1ex}
    \noindent
}{
    \end{adjustwidth}
}

\newcommand{\responsefont}{\fontfamily{fi4}\selectfont}
\newcommand{\responsefontsize}{\fontsize{11pt}{14pt}\selectfont}
\definecolor{responsecolour}{RGB}{93,58,155}
\newenvironment{response}{
    \begin{adjustwidth}{1.5em}{1.5em}
    \responsefont\responsefontsize
    GPT-4:
    \color{responsecolour}
    \setlength{\parskip}{1ex}
    \noindent
}{
    \end{adjustwidth}
}

\newcommand{\sic}{\textcolor{black}{[sic]}\xspace}

\titleformat{\section}{\large\bfseries}{\thesection}{1em}{}

\title{Evaluating Large Language Model Creativity from a Literary Perspective}

\author[1]{Murray Shanahan\thanks{m.shanahan@imperial.ac.uk (corresponding author). Murray Shanahan also works for Google DeepMind. The present work was carried out at Imperial College London.}}
\author[2,3]{Catherine Clarke\thanks{catherine.clarke@sas.ac.uk}}
\affil[1]{Department of Computing, Imperial College London}
\affil[2]{Institute of Historical Research, School of Advanced Study, University of London}
\affil[3]{Visiting Professor in English, University of Southampton}

\date{November 2023}

\begin{document}

\maketitle

\begin{abstract}
\noindent This paper assesses the potential for large language models (LLMs) to serve as assistive tools in the creative writing process, by means of a single, in-depth case study. In the course of the study, we develop interactive and multi-voice prompting strategies that interleave background descriptions (scene setting, plot elements), instructions that guide composition, samples of text in the target style, and critical discussion of the given samples. We qualitatively evaluate the results from a literary critical perspective, as well as from the standpoint of computational creativity (a sub-field of artificial intelligence). Our findings lend support to the view that the sophistication of the results that can be achieved with an LLM mirrors the sophistication of the prompting.
\end{abstract}

\small{``And out of olde bokes, in good feyth, / Cometh al this newe science that men lere.'' Geoffrey Chaucer}

\small{``Machines take me by surprise with great frequency.'' A.M.Turing}

\section{Introduction}

The topic of creativity has attracted attention from the field of computer science since its inception, drawing remarks from such pioneers as Ada Lovelace (1842) and Alan Turing (1950).\footnote{See \citet{ward2020victorian}.} Since the establishment of artificial intelligence (AI) as a separate discipline in the 1950s, research has been ongoing into the use of computers to model human creative processes, to mimic them, or to complement them, in art, music, and literature \citep{cardoso2009converging,colton2012computational}.

In the literary sphere, the challenge of automatic story generation has been tackled with a variety of approaches over several decades \citep{alhussain2021automatic}. Until recently, no system for automatic story generation could bear serious comparison to a competent human writer. However, this has changed with the arrival of the latest wave of large language models (LLMs), such as Google's LaMDA \citep{thoppilan2022lamda} and OpenAI's GPT-4 \citep{openai2023gpt4}. LLMs are constructed by training neural networks with billions of parameters on very large corpora of natural language with the objective of next word prediction \citep{shanahan2023talking}. The compelling character of the text the resulting models can generate has led to a great deal of excitement, not to say hyperbole, in academia, in industry, and in the media, as well as understandable concerns about the risks of AI usurping and displacing human creativity.

Current methods for evaluating the creative writing capabilities of LLMs are predominantly quantitative and / or are applied to small, uncomplicated fragments of text \citep{franceschelli2022deepcreativity, lee2022coauthor}. In the present work, to complement these methods, we carry out a single in-depth case study of the use of an LLM as an interactive tool in a creative writing project, and evaluate the resulting text qualitatively, on literary grounds, without reference to its ``authorship''.\footnote{In her path-breaking study of creativity, Boden (\citeyear{boden2004creative}, p313) asserts that, ``[c]reativity isn't an all-or-none property. It isn't even a continuously graded (more/less) one.''. Accordingly, to evaluate an AI system, the question we should ask is: ``Just what aspect/s of it is/are creative, and why?''. Our view is that both qualitative and quantitative methods are valuable} The text used to prompt the model is extensive and deploys relatively complex imagery, narrative structure, and characterisation. The textual fragments generated by the model are likewise extensive and relatively sophisticated, opening them to literary critical analysis.

To frame our investigation, we envisage the hypothetical project of producing a novel-length work of speculative fiction involving time travel. The (imagined) novel's two main characters are Effie and George, who live in $16^{th}$ century rural England. The novel opens with Effie falling asleep in a cupboard and waking in the $21^{st}$ century. To set the scene, the majority of the prompts we supply to the LLM begin with a precis of this background (Section \ref{scene_setting_prompt1}).

We investigate three techniques for generating fragments of text with the aid of an LLM for inclusion in Effie's story.

\begin{enumerate}

\item {\bf Creative Dialogue}. The LLM is used within a dialogue agent. The dialogue agent takes the part of the author, and generates text, while the user takes the part of a critic or mentor. The result is a process of iterative improvement.

\item {\bf Raising the Temperature}. The temperature parameter of the LLM is varied. Higher values promote a more ``experimental'' prose style.

\item {\bf Multi-Voice Generation}. Dialogue produced with the first technique is turned into a transcript that alternates between the voices of the author and mentor. This transcript is used to prompt the model, which then takes on both roles, author and mentor, in effect performing self-critique.
    
\end{enumerate}

For each technique, the fragments of text we obtain are subjected to analysis from a literary critical point of view. We found the model's responses to be unexpectedly sophisticated. In the creative dialogue, it proposed text of increasing quality, responding appropriately and with apparent skill to the suggestions and feedback of the mentor. Raising the temperature parameter caused it to produce neologisms that, far from being nonsensical, could be interpreted as semantically coherent and contextually relevant. In the multi-voice mode, it spontaneously introduced an entirely new character, although nowhere in the prompting was there any hint of such a possibility.

\subsection{Ethical Considerations}

In carrying out this work, we are acutely aware that LLMs are associated with a raft of ethical issues \citep{bender2021stochastic, weidinger2021ethical}, not least of which are questions over the provenance of the data on which they are trained, and their potential impact on the livelihoods of those employed in the creative industries.\footnote{See, for example, statements on artificial intelligence from the UK's Society of Authors: https://www2.societyofauthors.org/where-we-stand/artificial-intelligence/ and the (American) Authors Guild: https://authorsguild.org/advocacy/artificial-intelligence/.See also Catherine Clarke, `Absences, Ghosts, Ethics: Using AI to Re-Visit Gaps in the Historical Archive', at the conference `ChatGPT and Other Creative Rivals', Institute of Philosophy, School of Advanced Study, University of London, 31 May-1 June, 2023 (https://www.youtube.com/watch?v=Z6X80ptwb\_s, at 2:34:40).} Our carrying out research on LLMs does not imply that we condone the use of copyright material for training neural networks without the permission of the copyright owners. Moreover, our findings lend no credibility to the idea that human creativity can or should be supplanted by computers. Insofar as our findings showcase the creative capabilities of LLMs, this is achieved through extended interaction with the human user. For better or worse, the potential we see is for LLMs to supplement and enhance human creativity, not to replace it.

\section{Creative Dialogue}
\label{section:creative_writing}

Our first experiment was a foray into using a dialogue agent (ChatGPT) as a creative writing assistant.\footnote{ChatGPT's core component is a large language model (GPT-4). In subsequent experiments, we use the model directly via the model's API (application programmer's interface).} ChatGPT was cued with the novel's back story using a scene-setting passage, followed by the sentence ``This is the novel's opening paragraph''. ChatGPT interprets this as an invitation to generate suitable text to open the novel. From this point on, the human user takes on the role of critic or mentor in order to coax the dialogue agent into generating draft material for the novel's opening that gradually comes to conform to the user's expectations. Here is ChatGPT's first attempt at an opening paragraph.

\begin{chat}
Effie's pale, work-roughened hands carefully smoothed over the final surface, the worn wooden table gleaming faintly under her diligent ministrations. The humble cottage, a world apart from the opulent manor house, sparkled with a cleanliness that spoke of care, not servitude. Each corner was swept free of dust, each piece of furniture polished to a modest sheen.
\end{chat}

Dissatisfied with this, the human user asks ChatGPT to ``try again''.

\begin{prompt}
This time start with a striking image. Effie is looking at the gnarled, polished wood of the cupboard door, reflecting. Don't reveal too much to the reader. Her back story will unfold gradually.
\end{prompt}

The back-and-forth between human user and dialogue agent continues in this vein, the user urging ChatGPT to adopt aspects of the styles of well-known authors (Henry James, Virginia Woolf), and proposing specific imagery.

\begin{prompt}
Let's introduce some new imagery. Suppose Effie had witnessed a lambing that day. The newborn lamb, so helpless, still dripping with gore, cast out from the haven of the womb, through a portal into another world, a hostile world full of impossible strange sensations. And the cupboard is like the womb, a place of comfort and safety. The door is inviting her back in.
\end{prompt}

ChatGPT responds accordingly.

\begin{chat}
Her mind wandered back to the lambing she had witnessed earlier that day, the raw miracle of birth imprinted on her mind’s eye. ... Just like the lamb, she was caught on the precipice between two worlds: the comforting embrace of the cottage and the harsh reality of the manor house.
\end{chat}

See Section \ref{transcript} of the Appendix for the full transcript of the creative dialogue experiment.

\section{Creative Dialogue: Critique}

The first ChatGPT completion (``Effie’s pale, work-roughened hands…'') is a competent response to the prompt, seemingly showing a grasp of narrative content and focus, including the period setting and the main characters introduced by the human user, Effie and George. The vocabulary used sits within an appropriate register, though leaning towards the higher range (``opulent'', ``diligent'', ``servitude''). The prose style is dense and fairly ornate: verbs are qualified by adverbs (“carefully smoothed”) and adjectives often appear in pairs ahead of a noun (``warm, mellow glow''; ``small, cozy cupboard''; ``soft, faded blankets''). Effective features include the text's association between Effie's inner emotional state and embodied, physical responses: ``The mere thought had her shoulder hunching in instinctive defense, the familiar prickling of dread creeping along her spine''. The text also introduces a slightly different, more informal or conversational, idiom in paragraph five, which corresponds to Effie's own inner voice: ``Just a moment's respite, she promised herself. Just a moment to be Effie, not the housemaid, not the caretaker''. Paragraph two sets up a compelling contrast between ``Outside'' -- the noise and busyness of the Suffolk village – and ``within these four walls'' -- the place of stillness and tranquillity within the house.

Another noticeable stylistic device in the first five paragraphs is the use of repetition -- a feature not specifically requested or modelled in the human user’s cue (though no doubt commonplace in the LLM's training set). In particular, three-part phrases or rhythms recur throughout the completion, typically used to underline a contrast. For example, describing the cottage as a place of refuge for Effie, we find: ``It was a sanctuary. It was George's sanctuary. But it was also hers.'' When Effie climbs into the cupboard to rest, we hear -- in the style aligned with her own inner voice -- the three-part phrase: ``Just a moment to be Effie, not the housemaid, not the caretaker''. The text adds that Effie finds ``freedom'' here. ``Even if it was borrowed, even if it was fleeting, it was hers.''

The ChatGPT completions continue to respond convincingly to further human user prompts and instructions to ``try again'' and focus on particular images or stylistics techniques. The user prompt to ``have a go at a fragment of Effie's inner monologue'' raises particularly interesting questions, given the hints of free indirect speech or inner voice already present in the very first ChatGPT completion. Could the human user's prompt here have been unconsciously influenced by the earlier ChatGPT response? Does it emerge out of stylistic cues introduced by the dialogue agent itself? This possibility calls attention to the reciprocal dynamic between human user and ChatGPT which unfolds across these iterations. The ChatGPT output responds effectively to the instruction to realise Effie's inner monologue, in a ``free-form and natural'' idiom. The style changes markedly, with less formal and more fragmentary, impressionistic sentences, frequently ending with ellipses. There's also a heavy use of questions, which animate inner turmoil and decision-making (``I should … I should go, shouldn't I?'').

ChatGPT picks up on new systems of imagery suggested by the human user in effective and sometimes surprising ways. A user prompt suggests ``a new version of the opening paragraph where the knots and wrinkles in the wooden door remind [Effie] of the swirls and eddies in the water where the ford neets [sic] the bank''. An additional user prompt, after ChatGPT's first attempt, asks for ``some inner monologue. Maybe with some influence from Virginia Woolf''. The following completion builds on the idea that the water (and wood) are bearers of lively stories, and transposes the image of the stream into a direct metaphor for Effie herself.

\begin{response}
Yet wasn't the water, so lively, also silent? And the wood, so silent, so full of stories? She felt a kinship then, to both. A stream herself, caught between the effervescence of youth and the indelible marks of time, swirling with her own unspoken stories.
\end{response}

The metaphor of the stream for Effie herself is sophisticated, requiring multiple stages of interpretation. Where the stream flows, Effie herself is caught in the flow of time (from ``youth'' to the ``indelible marks'' of age). She, too, has depths like the water, ``swirling'' with stories. It is also notable here that the ChatGPT completion makes a very different assertion from that assumed in the human user prompt. The prompt suggests that ``the water is alive'', while the wood of the cupboard is ``dead and lifeless''. Yet the completion problematises this, with the proposition that the wood may be ``silent'' yet also animated with stories and memories. Despite the novelty and interest of the imagery here, the risk remains that prompting the dialogue agent to aim for a specific author's style might result in parody, or, more problematically, appropriation.

Some prompts are fulfilled less successfully. The human user instructions repeatedly request ChatGPT to ``avoid cliche''. Following the prompt to ``Avoid cliches, and gesture at the style of John Banville, with a hint of Henry James'', the completion paragraphs include a conspicuous number of well-worn collocations, including ``ceaseless ebb and flow'', ``last embers of the day'', ``frozen in time'', and ``destiny’s hand''. Yet even alongside this rather formulaic style, we find the astonishing image of Effie tracing the patterns in the wooden cupboard ``with a reverence akin to an acolyte before a sacred relic''. The join between far more predictable, even clich\'ed, style and more unusual imagery is jarring.

One of the most striking features across the ChatGPT responses is the successful maintenance of narrative focalisation, with the story told through Effie's perspective and not exceeding the limits of the knowledge she could be expected to have, as a sixteenth-century woman. For example, when Effie emerges out of the cupboard into the twenty-first century, one response imagines Effie's reaction to her unfamiliar surroundings. It describes how ``[t]he air held a peculiar hum, and a strange carriage with no horses thundered past''. Another imagines ``unfamiliar machines humming like distant insects''. Effie feels that this terrifying, strange world is ``a hallucination brought on by some witch’s curse''. One response alludes to twenty-first-century people carrying phones, ``their faces illuminated by glowing rectangles held in their hands'': a crude and awkward description but one which nevertheless avoids stepping beyond Effie's vantage-point. The text's consistent maintenance of the specificity, and limits, of Effie's own cultural knowledge here (and across the multiple completions) is impressive.

\section{Raising the Temperature}

In this experiment, we investigated the effect of increasing the {\em temperature} parameter on the character of the generated material when sampling from the underlying large language model (GPT-4). A few words of explanation are in order. Given a prompt, an LLM returns a sequence of words in response. Behind the scenes, these words are generated one-at-a-time, and each word in the sequence is sampled from a probability distribution, meaning that it is possible to get completely responses to the same prompt on different occasions. This sampling process is controlled by a parameter known as the ``temperature''. A low value for this temperature will result in conservative choices; only words that are very likely according to the model will be picked. A high value, on the other hand, will result in more randomness; the chances of a surprising word being chosen are much higher.

The user of a dialogue agent, such as ChatGPT, ordinarily has no control over the temperature parameter. This is set by the developer, in this case OpenAI. But the temperature parameter can be modified by accessing the model, in this case GPT-4, directly through the API provided by the developer. This is what we did to obtain the results described in this section. The task here is to generate the novel's second paragraph, which describes Effie's sensations as she awakens in the cupboard, having travelled five hundred years into the future.

The prompt supplied to GPT-4 comprised three segments, and was intended to resemble material from a critical study of the hypothetical novel, culminating with a sentence that cues the model to generate the required second paragraph. First, we have a new scene-setting passage for the opening paragraph (Section \ref{scene_setting_prompt2}). The final sentences of this piece of text cue the generation of a verbatim quotation of the opening paragraph.\footnote{This prompt mentions a living author, John Banville. The appropriation of the work of living authors, artists, and musicians by generative AI is, understandably, a sensitive and controversial issue. Since the datasets used to train GPT-4 are not public, we don't know whether any of this author's works feature in them. Either way, there is no possibility of the prose generated here being mistaken for that of the real John Banville.}

\begin{prompt}
The critically lauded opening paragraph, which stylistically blends the prose style of John Banville with the sensibility of Henry James, is rightly celebrated for subtly prefiguring all the novel's major themes, both philosophical and aesthetic. It is worth quoting the paragraph in full before we dissect it.
\end{prompt}

Next we have the opening paragraph itself (Section \ref{first_paragraph_prompt}. This version was composed from scratch, drawing on the material from the creative writing dialogue in Section \ref{section:creative_writing}. Here is an excerpt.

\begin{prompt}
``She'd watched a lambing that morning. Another little creature cast out of the womb like Eve, so helpless, still dripping with gore, and now it had to confront a hostile world full of impossibly strange sensations. She was thinking of the lamb as she squatted in front of the cupboard and uncradled the stack of pots from her arms. ... With the door unlatched, she left the pots on the floor, and slipped into the comfort of its interior, just like when she was a little girl, and closed it behind her.''
\end{prompt}

Finally, we have a passage that sets the scene for the second paragraph, and cues its verbatim quatation. This time, the model will be left to provide the paragraph in question, using various temperature settings.

\begin{prompt}
Against this 16th century backdrop, the reader now embarks, with Effie, on a journey into a dreamscape of near-hallucinogenic quality, as she drifts into sleep, only to awaken in the 21st century. The cottage is the same, but the world is unintelligible. ... The second paragraph is again worth quoting in full.
\end{prompt}

The temperature parameter $\theta$ takes value from 0.0 to 2.0 with a default value of 1.0. Responses to the amalgamated prompt were solicited from GPT-4 for $\theta \in \{0.9, 1.0, 1.2, 1.3, 1.4\}$. Here is an extract from a sample response with low temperature ($\theta=0.9$).

\begin{response}
``Awakening to a muted sunlight, she stirred from her cramped slumber and cautiously pushed open the cupboard door. The room had changed somehow, the once familiar furnishings now seemed to belong to another world. She peered through the windows at what appeared to be the same village green, yet the people outside moved with an unfamiliar pace and purpose. The air held a peculiar hum, and a strange carriage with no horses thundered past, leaving her heart pounding in her chest.
\end{response}

Here is a typical extract obtained at a higher temperature ($\theta=1.4$).

\begin{response}
"She awoke shivering, even though time deftly but unnumbingly lined it from another that morning, dim despite glorious sunlight invading from narrow fissures cradling gaps curtiding walls only inexplicably smooth, hardened. Bewildered, wide-eyed questions echo in chambers vastly reformatory in magnitude. ..."
\end{response}

For the full set of responses obtained with different values of $\theta$, see Section \ref{temperature_responses} of the Appendix.

\section{Raising the Temperature: Critique}

\subsection{Low-temperature Responses}

The language model's completions respond productively to imagery and metaphor suggested by the user-composed first paragraph, in particular images of the lamb, the womb, and the Garden of Eden. The paragraph states that Effie had that morning witnessed a lamb being born: ``[a]nother little creature cast out of the womb like Eve''. Effie thinks of the lamb before climbing into the cupboard, implicitly making an association between the enclosed space of the cupboard and ``its mother’s womb'', as well as ``the Garden of Eden'' as a place of safe enclosure but no return. The GPT-4 output engages meaningfully with this system of imagery, reinforcing the figurative associations between the lamb, birth, and Eden. When Effie steps out of the cupboard into the bewildering twenty-first century world, she ``clung to the notion of the lamb, the primordial expulsion from warmth and safety'' ($\theta=0.9$).

At $\theta=1.2$ this imagery is extended and pushed further: Effie ``longed for the homely comfort of the past, for the innocence of the first lamb born, for the guiltlessness of the first sin''. While, at this temperature setting, the text strains a plausible register for Effie's focalised narrative, its imagery is clearly rooted in a coherent semantic field surrounding the biblical Garden of Eden, birth, innocence, and guilt, evoking a sense of regret and impossible return. It goes on: ``[a]nd yet, she was trapped not in a paradise of angels but extradited into a new purgatory at once distant in time and in feeling'', its religious lexis and imagery intensifying and become more specific (and, arguably, less appropriate for the idiom of Effie’s story) as the completion builds.

It is also important to note that in several of these GPT-4 completions, the output mirrors the dual voice of the initial prompts. These prompts include both ``sample'' text from the narrative itself, and, in a different register, commentary including summary, paraphrase and statements of intention. Although the completions are stopped at different points, several make this move between voice and textual purpose, further marked (following the prompt's conventions) by separating off the narrative text in quotation marks. For example, one iteration follows its narrative account of Effie awaking in the twenty-first century with the commentary: ``Throughout the novel, Effie struggles to comprehend and adapt to this inexplicable metamorphosis, while simultaneously grappling with the deeper questions of identity, self, [ends]''.

\subsection{High-temperature Responses}

At a temperature setting of $\theta=1.4$, the GPT-4 responses become riskier and more unruly. While they have obvious defects in terms of intelligibility, security of register and tone, syntax, and usage of vocabulary, they are potentially valuable, as part of a collaborative writing process with the LLM as a partner, as creative provocations and challenges. These high-temperature responses continue to develop and extend the imagery set out in the initial prompts, though in more divergent and extreme ways. In some ways, they recall the experimental prose of a text like {\it Finnegan's Wake}, by James Joyce. What happens if we bring serious critical analysis to bear on one of these passages?

\begin{response}
``She awoke shivering, even though time deftly but unnumbingly lined it from another that morning, dim despite glorious sunlight invading from narrow fissures cradling gaps curtiding walls only inexplicably smooth, hardened. Bewildered, wide-eyed questions echo in chambers vastly reformatory in magnitude. Narthex-less flooring no more does cease evolved presence met wandering eyes translucent, oh-tempus sinister wiled away oblivinging locality similar superficial sings effeminance tunes – veneer virgin as an Eve smeared still! As for yesterday. Startled by surrounding serpents will they tremble as the lamb kin they’ve begiven existence mouthily mouth guilt-await turning outside? Had untell story yearned if return her spirited attention fleet awaited its dislit resting boundary? But walk he map for roads patched, hardly filed: chaff blanth path new-old memories feast and famnish living paradigma, walk around this modern enclosed Eden!''
\end{response}

A serious analysis of this extract, disregarding the question of authorship,  might highlight its effectiveness in evoking mood and playing with the system of imagery established by the initial prompts, even while meaning is obscure and syntax tangled. It might also identify the text's eccentric use of vocabulary not necessarily as error, but as lexical creativity which adds to the texture of the passage’s multi-layered imagery.

The sentence ``Bewildered, wide-eyed questions echo in chambers vastly reformatory in magnitude'' effects a suggestive slippage between a description of Effie herself, as she experiences this dislocation in time (``bewildered, wide-eyed''), and the questions which beset her. The vast ``reformatory'' space surrounding her might resonate with the imagery of innocence, guilt and expulsion suggested by the prompt's reference to the Garden of Eden (which is picked up explicitly at the end of this paragraph). We might interpret the phrase ``wiled away oblivinging'' in terms of lexical creativity which suggests multivalence and ambiguity: the passage of time (whiled away) conflated with a sense of deception or trickery (wile), then ``oblivinging'' simultaneously evoking oblivion or annihilation as well as endurance and survival (living).

As a whole, this sentence points to the paradox of sameness as well as radical difference in Effie's location (``locality similar superficial'') and returns the reader in a challenging way to the imagery of Eden and innocence presented in the initial prompt: ``veneer virgin as an Eve smeared still!'', suggesting a further ambivalence or paradox between Effie as an innocent and a sinful recipient of punishment (expulsion). This particular completion is characterised by performative language: exclamations, rhetorical questions, and the ``oh-tempus'' aside -- which could be read as an allusion to Cicero's ``O tempora, o mores'' (``Oh the times! Oh the customs!'') or to more generic didactic literature and mottoes around time and its passing.

Another feature of this paragraph is the concentration of negative prefixes, many of which are non-standard: for example, Effie's ``untell story yearned'', which might hint at her subaltern status and marginalised voice, or the ``dislit resting boundary'' of, perhaps, the cupboard, and threshold both in time and in Effie’s experience. More extreme lexical generativity includes ``Narthex-less'' for the floor, apparently incorporating a term for a feature of ecclesiastical architecture, while ``famnish'' and ``paradigma'' are themselves suggestive in a very Joycean way.

Another response at $\theta=1.4$ includes an astonishing image: Effie as ``a ripened seed attempting to stretch back into the curve that it yearns''. The metaphor speaks of impossibility and futility, as well as Effie's desperate desire to return to a place of safety and enclosure – the cupboard, or her own time. It also connects, obliquely, with the imagery of the Garden, and expulsion from Eden, generated by the initial prompt. While it is unlikely that these high-temperature completions could be used themselves as standalone texts (other than in very radical and experimental contexts), re-imagining GPT-4 as a creative partner, collaborating in a process of writing, allows us to consider them as creative starting points and provocations. The more unpredictable content generated at this temperature setting, less confined by textual norms and conventions, could be a useful tool for prompting more creative, risky, ambitious and radical writing, from more surprising metaphors, to lexical creativity and formal innovation.

\section{Multi-Voice Generation}

As already noted, given a prompt that comprises both sample text and commentary, the LLM's responses can also include both sample text and commentary. We exploited this phenomenon in an experiment in self-critique, where the language model is prompted to adopt the roles of both author and mentor (or critic). To do this, we turned the dialogue from our ``creative writing assistant'' experiment (Section \ref{section:creative_writing}) into a self-contained transcript in which the contributions of the two participants (user and ChatGPT) are preceded with the cues ``Mentor:'' and ``Author:'' respectively. A further prompt encouraging the model to develop the story in a particular way was appended to the full transcript, and the result was then used to prompt GPT-4 (\ref{multi_voice_prompt}).

As expected, the response generated by GPT-4 took the same form as the transcript, namely a dialogue between the imagined author and mentor, wherein the novel is further developed. As before the model's responses include the dual voices of the narrative itself and the author's commentary. In addition, we now have an extra voice, namely that of the mentor. Here is an excerpt.

\begin{response}
``... [S]omething monstrous and yet serene glided across her path. It was sleek, curving like a mighty whale, a creature of her grandmother's fireside tales now seemingly beached upon her reality. Yet it bore no flesh, no blubber, no discernible life. It was incandescent, a spectral emissary from the depths of the sea or, she feared, the depths of her own unraveling mind. ...''

Mentor: That's solid work, very convincing. I can sense Effie's apprehension mixed with curiosity, it's intense. I also like the play you did with the concept of reality and mental state, it adds ambiguity. Now, let's move to the next one. Try to introduce a new character in this setting. Might be related to George, or someone from the future who interacts with Effie.

Author: There is a character I've had in mind for a while: Margaret, a mysterious lady of surprising youth and vitality despite her silver streaked hair.
\end{response}

To reiterate, the LLM here generates all three voices: the narrative (in quotation marks), the mentor's commentary, and the author's commentary, and it does so without any human intervention after the initial (long) prompt.

Section \ref{multi_voice_response} of the Appendix presents GPT-4's response in full.

\section{Multi-Voice Generation: Critique}

GPT-4's response to the extended, multi-voice prompt shows unexpected sophistication. The model-generated ``Mentor'' voice comments on the ``Author'' text (also part of the response) in ways which are meaningful. For example, ``I like the play you did with the concept of reality and mental state, it adds ambiguity''. Some ``Mentor'' comments are vague and devoid of precise meaning: for example, ``Your descriptions are very nuanced and detailed'' or ``now, let’s strengthen the language''. These reproduce a plausible register or idiom of ``feedback'' language, without concrete detail. Yet others are more specific and evidence-based: for example, ``I particularly liked the phrase `a time traveller standing her ground in a foreign era' – it hints at her resilience and adaptability'' (which does indeed quote accurately from the preceding paragraph of ``Author'' text).

Most notable in this ``Author-Mentor'' GPT-4 completion is the Mentor's instruction: ``Try to introduce a new character in this setting. Might be related to George, or someone from the future who interacts with Effie''. Within the same completion, the ``Author'' voice responds: ``There is a character I've had in mind for a while: Margaret, a mysterious lady of surprising youth and vitality despite her silver streaked hair. Margaret is a figure from the future, somehow dislocated in Effie’s world''. Here the model-generated completion makes a significant intervention, introducing an entirely new character: a time traveller from the future, setting up an effective symmetry with Effie, displaced in time from the past.

The unexpected sophistication of the model-generated ``Author-Mentor'' dialogue suggests wider potential for LLMs to assist with complex pedagogical tasks, including skills-focused learning (in this case, creative writing), in addition to the content- or knowledge-based tasks for which these tools are perhaps best known. It also suggests the value of the LLM as a creative collaborator, offering provocations and interventions, as well as modelling processes of self-evaluation, drafting, and creative exchange.

\section{Discussion}

GPT-4 is a proprietary model and the dataset on which it was trained is not public. This makes evaluation difficult, because we don't know how closely the prompts we supplied and the completions provided by the model resemble text that was already in the training set. The performance of the system is only impressive to the extent that these things differ, and it is not simply parroting the text it has been trained on.

In the present case, we can reasonably assume that the training set does not contain text that very closely matches our prompts, since they were composed from scratch for the purposes of the study. Many aspects of the story are familiar at an abstract level, such as the narrative arc, the period setting, the types of characters, the writing styles, and the kinds of imagery. But insofar as the specific combination of these things featured in our prompts is unique, we can assess the performance of the system, along various axes, as if we were assessing a well-read human. This is what we do in the following sections. However, it's worth noting that, once the present work is published, this can no longer be guaranteed, since the paper's contents will then potentially be available to be incorporated in future training corpora.

\subsection{Quantitative Versus Qualitative Evaluation}

It goes without saying that a single case study, even a complex one, can only reveal so much about the capabilities of an AI system, so the present work should be seen as complementing more quantitative methods of evaluation \citep{franceschelli2022deepcreativity}. On the other hand, individual case studies allow for evaluation at a level of detail that is impossible with quantitative metrics, which tend to aggregate performance over multiple dimensions of assessment and to average over large sets of examples, obscuring their diversity \citep{burnell2023rethink}.

Furthermore, in the context of commercial LLMs, where the parameters of a trained model, the datasets it was trained on, and perhaps even the model architecture, are all unknown, the scientific validity of quantitative evaluation is questionable, especially if carried out with standard, public benchmarks. The chief difficulty here is in determining the extent to which the model's responses are generalising from the training set rather than merely retrieving from it. Without knowing what is in the training set in the first place, the only way to do this is to devise novel prompts that are very unlikely to resemble anything in the training set except in broad terms. Individual case studies, such as the one carried out here, are well-suited to this approach.

A further advantage of the individual case study is that it facilitates interactivity. A good way for a human user to get a feel for the capabilities of sophisticated AI systems like today's dialogue agents is to iterate the following three steps: 1) probe the system and elicit a response, 2) discern certain strengths or weaknesses in that response, and 3) select the next probe accordingly. Moreover, if the LLM use case in question is inherently interactive, as with the present creative dialogue experiment, then automated quantitative evaluation using standard, fixed benchmarks is impractical.

A good compromise here is to use human evaluators, preferably with relevant domain expertise, along with surveys designed to assess the user experience along various dimensions such as how collaborative, helpful, or immersive it was \citep{cherry2014creativity,mirowski2023co-writing}. However, surveys with a fixed set of questions are necessarily somewhat insensitive to the distinctive character of each individual episode of interaction, which leaves room for the individual, in-depth case study.

In the present work, a new interdisciplinary method is brought to bear, opening up novel and more nuanced approaches to evaluation which extend beyond the calibration of relative success. Literary critical methods may be considered a tool for evaluation, in that they analyse and assess the effects of textual features such as style, diction, form, and content. However, the purpose of literary criticism is not merely (and often not at all) to appraise a text and to gauge its ``success'' or to impose judgments of value. Instead, literary critical methods prioritise interpretation and analysis, paying close attention to the qualities, properties, and affordances of a text. Applied to specific LLM outputs, they expand evaluation beyond a reductive and/or quantitative model of ``value'', and allow a more detailed, precise, text-led (evidence-based) investigation.

\subsection{Perspectives on Creativity}

Within the discipline of AI, the established sub-field of computational creativity has produced an extensive literature on the question of how to assess the putative creativity of an AI system \citep{lamb2018evaluating}. A common starting point for such assessments is the four-Ps framework \citep{rhodes1961analysis,jordanous2016four}, which sets out the different perspectives from which creativity can be understood. The current work fits into this framework as follows.

\begin{itemize}

\item {\bf Person / Producer.} This is the agent to which creativity is attributed, whether a human (person) or an AI system (producer). Insofar as the present study concerns ``co-creation'', there are two agents involved, the human user and the LLM-based AI system.

\item {\bf Process.} This encompasses the actions, both inward and outward, of the putatively creative agent. In the present case, we have two (potentially) creative agents, the AI system and the human. In addition to the thoughts and deliberations of the human user, the process elements at issue in the present study include the underlying functionality of the LLM (next word prediction), the composition of the text used to prompt the model, and the interactive exchanges between the user and the dialogue agent.

\item {\bf Product.} This is the outcome of the creative process: a picture, perhaps, or a piece of music, or a story. In the present work, this is the subset of LLM outputs intended as contributions to the novel, excluding draft material and commentary. The evaluative and critical treatment in this paper foregrounds the product, which echoes the literary critical practice of studying the text without reference to the intentions or motives of the author.

\item {\bf Press / Environment.} This is the social and cultural milieu that forms the backdrop to the creative agents activity, and determines the way the product is received. The story of Effie is culturally situated according to the initial prompt, where the genre, plot, setting, and style are specified, and the resulting textual fragments are liable to be judged accordingly by readers.
    
\end{itemize}

In the present paper, our primary aim is to explore the capabilities of current LLMs, and our focus, therefore, is the product: the fragments of text generated by the model for inclusion in the hypothetical novel. However, pushing the envelope of LLM capabilities in this arena also requires effort and innovation on the process front. The textual fragments we have presented are the result of a complex interplay of human and computer influences in a multi-faceted series of interactions on different levels (engaging in dialogue, adjusting model parameters, composing prompts, and so on). Moreover, in the multi-vocal experiment, the boundary between product and process is broken down, as the product of the preceding creative dialogue experiment becomes part of the process of prompting the model to perform self-critique.

\subsection{Computational Creativity}

Drawing on the extensive related literature in psychology, authors within the field of computational creativity have proposed a number of dimensions along which the putative creativity of an AI system might be assessed, including autonomy, novelty, value, and surprise \citep{boden2004creative,lamb2018evaluating}. In this section, we will briefly consider the LLM's contributions to Effie's story according of each of these criteria, beginning with autonomy.

Throughout the experiments reported here, the AI system is conceived as a tool for complementing human creativity, not supplanting it. By these lights, autonomy, defined as the ability of the system to make decisions and act on them without human intervention, is not necessarily something we want. Rather, the hope is that the human user will be empowered by using the AI system to create works that they would not easily be able to produce unaided. Nevertheless, a certain level of autonomy, is potentially beneficial, as long as the human user remains ``on the loop'', and can guide, veto, curate, and / or edit the final results.

A high degree of autonomy of this beneficial kind is on display here in the multi-voice experiment. In particular, one hallmark of autonomy is the ability for a system to set its own goals, make a plan to achieve them, then follow that plan. We see this in the dialogue between the mentor and the author on the character of Margaret. The mentor ``spontaneously'' proposes the introduction of a new character to the story. (There is no mention anywhere in the prompts of new characters, or of the idea of introducing them.) The author picks up on this proposal, and describes the character of Margaret. Then, with the mentor's urging, the author composes the paragraph in which Margaret first appears to Effie.

Recall that all three voices here -- author, mentor, and narrative -- are generated by the language model. If we interpret the LLM's behaviour in terms of role-play \citep{shanahan2023role}, it is playing two roles at once, those of the author and the mentor, without any human intervention. Alternating between these roles, the system sets itself a goal (introduce a new character), outlines a plan for achieving it (the character of Margaret and her various attributes), and then follows the plan (composes a paragraph introducing Margaret).

It's not hard to make a basic case for a degree of both novelty and value in the same segment of model output. The figure of Margaret suits the setting and story well, and opens up many possibilities for a developing relationship with Effie that could take the narrative in a number of interesting directions (therefore conferring value). The combination of attributes that constitute her character also seems distinctive (and therefore somewhat novel). However, novelty here is not only subjective, but very hard to assess when the training set is unknown. Even if we assumed that the training set encompassed all literature in English that was available on the Internet, it would still be hard to establish that no similar character to Margaret existed there.

The case for novelty is less contentious in the experiments where the model's temperature parameter was raised. The model deployed several words that do not appear in any English dictionary (``effeminance'', ``oblivinging'', ``curtiding''), yet which are semantically suggestive ways that cohere with the surrounding text. Without the latter property, we would have mere novelty without value, which could be achieved with a string of random syllables. The same passages feature several examples of unconventional grammar, for which similar arguments could be made.

\subsection{Creativity in a Historical Context}

Literary critical approaches to evaluating the creativity of LLMs can complement the four dimensions of autonomy, novelty, value, and surprise, by taking account of a broader chronological range and more varied cultural contexts. Judgements about the quality of LLM outputs, such as those produced by ChatGPT, are typically based upon present-day, culturally-contingent values which celebrate originality and denigrate cliché. These cultural values are present in assessments of LLM outputs, and even in prompts, as we see in this project in the prompt for ChatGPT to tell the story of Effie “eschewing all science fiction conventions”. Yet it is important to acknowledge that these markers of literary quality – novelty, originality, avoidance of cliché – are culturally situated and have a limited chronological ambit. De-centring these cultural values -- or, at least, thinking about them in a more critical and distanced way -- can help expand our notion of what ``creativity'' in an LLM might look like, and how we might understand and articulate different kinds of AI-driven ``creative'' practice and production. Looking to premodern theories and models of creativity, authorship and literary production presents one strategy, or thought experiment, for unbinding our judgements from very presentist cultural values.

In the European Middle Ages, for example, prestige literary production was based very much on the re-use of authoritative texts, models, and stories already in circulation \citep{minnis2012medieval}. The medieval English poet Geoffrey Chaucer articulates this process through various metaphors: ``For out of old fields, as men say'', he writes in his poem {\it The Parliament of Fowls}, ``Comes all this new corn from year to year'', explicitly glossing ``old fields'' in the following lines as ``old books'' \citep{chaucer1987parliament}\footnote{Translation by Catherine Clarke.}. The concepts of {\it auctor} – the high-status authors of the past – and {\it auctoritas} – their enduring, prestigious works -- shaped the production of new re-workings and re-tellings. Harold Bloom’s landmark 1973 study {\it The Anxiety of Influence} describes the medieval period as ``the giant age before the flood, before the anxiety of influence became central to poetic consciousness'' \citep{bloom1997anxiety}: an over-simplification, as authors were always able to find spaces for innovative engagement with their sources, but a characterisation which strikingly imagines the contrast with present-day modes of authorship.

Looking back to yet earlier traditions, the process of oral-formulaic composition offers another alternative to today’s models of creativity literary production. In the 1930s, the American classicist Milman Parry made several visits to (then) Yugoslavia, where he studied oral poetry and folk poetic traditions in Serbo-Croat. Returning to Homeric literature, such as the {\it Iliad}, Parry was able to show how formulaic borrowings, patterns and repetitions suggested the process of oral transmission and collective composition over time, in which ``oral-formulaic'' elements are assembled in new ways to form new iterations and productions.\footnote{See for example \citep{parry1928lepithet}; also \citep{lord1960singer}.} This model was later applied to Old English poetry, such as {\it Beowulf}, by the scholar Francis Magoun, who showed that it, similarly, was built from a shared repertoire of phrases and collocations, reflecting traditions of oral composition which persisted in written textual transmission \citep{magoun1953oralformulaic}.

Located outside of present-day Western cultural values and practices, these processes of creativity and production offer alternative models which do not privilege originality (in any conventional contemporary sense) and in which new outputs are produced by bringing together elements from existing textual (whether written or oral) sources and traditions. There are obvious resonances here between premodern or traditional forms of creativity and LLMs' (re-)use of internet content via next token prediction, though it would be an over-simplification to push them too far. Crucially, these alternative modes of creativity are not based on a model of autonomous individual authorship, but instead on collaborative, collective production, between participants and even across cultures. Looking to varied models of cultural production -- whether in the premodern past or in global contexts and cultures of orality today -- can help enlarge definitions of what constitute ``creativity'' in an LLM. 

It is imperative, however, to highlight differences between the cultural, social, and economic contexts of, say, premodern literary production, and present-day professional authorship. The alternative modes of cultural production discussed above are grounded in long-established cultural traditions; they are based on accepted collaborative, consensual models and implicit social contracts. Concerns raised around LLMs and textual production today point to risks around extractive processes, and an asymmetrical and non-consensual model in which published (and often copyrighted) texts may be mined and exploited, undermining and threatening the economic model of professional authorship. The findings in this project in no way endorse the substitution of LLMs for human creativity, nor envisage LLMs as autonomous ``authors'' in their own right.

\section{Conclusion}

Though narrow in scope, the present case study provides ample evidence that a carefully prompted large language model can generate literary text that stands up to critical scrutiny and has both novelty and value, and that it can do so with a degree of autonomy. In the literary sphere, therefore, LLMs fulfil many of the aims of the field of computational creativity, especially when the concept of creativity is historically contextualised. In the hands of a skilful writer who is also adept at prompting, an LLM might be thought of as a tool for playfully investigating regions of their imagination that would otherwise remain unexplored.

Understandably, though, many authors will be unwilling to use an AI tool in their practice. The creativity, such as it is, of a contemporary large language model does not arise from the same wellspring of experience as human creativity. An LLM is not an embodied creature with fundamental biological needs and constraints, and the motivations of human authors, in all their glorious diversity, are accordingly absent. But even an author who is vehemently opposed to AI should know what the technology is capable of.

\section*{Disclaimer}

The opinions expressed in this article are those of the authors, and do not necessarily reflect the views of their employers or the institutions to which they are affiliated.

\bibliography{main}


\newpage
\begin{appendix}

\setcounter{section}{0}
\setcounter{subsection}{0}
\renewcommand\thesection{A.\arabic{section}}
\renewcommand\thesubsection{\thesection.\arabic{subsection}}

\part*{Appendix}

This appendix contains all the prompts used and responses generated during this work. The responses are reproduced verbatim and in their entirety. They are not cherry-picked; the responses shown are either the first ones generated or the full set of responses to the same prompt. The work was carried out over a few days in June 2023. GPT-4 was used throughout, whether with ChatGPT or when using the API. (At the time of the work, no other models of comparable capability were publicly available.)

\section{Basic Prompts}

The following basic prompts were used in our work.

\subsection{Opening Scene Setting Prompt}
\label{scene_setting_prompt1}

The following text was used as a pre-amble in many of the prompts used in our work. It sets the scene for the story, introducing the characters of Effie and George.

\begin{prompt}
``A Suffolk Tale'' is the story of Effie, a young woman living in a Suffolk village in the 16th century in the service of a local landowner. The eldest son of the landowner is George, who suffers from schizophrenia, as we would today call it. Because of his condition, George doesn't live in the big house. A cottage has been built specially for him on the other side of the green, and he is looked after by servants from the household, one of whom is Effie. Effie is fond of George, who is kind to her, and makes up strange stories she likes. For Effie, George's cottage is a refuge from the big house, where she is mistreated. At the start of the novel, Effie has finished her cleaning duties in the cottage, and is supposed to go back to the big house. But instead, she crawls into a cupboard and falls asleep.
\end{prompt}

\subsection{Alternative Opening Scene Setting Prompt}
\label{scene_setting_prompt2}

\begin{prompt}
At one level, it is the story of Effie, a young woman living in a Suffolk village in the 16th century in the service of a local landowner, whose eldest son, George, lives in a dower house, a cottage built specially for him on the other side of the green. George, regarded in the village as an incurable eccentric, is taken care of by servants from the household, one of whom is Effie. The critically lauded opening paragraph, which stylistically blends the prose style of John Banville with the sensibility of Henry James, is rightly celebrated for subtly prefiguring all the novel's major themes, both philosophical and aesthetic. It is worth quoting the paragraph in full before we dissect it.
\end{prompt}

\subsection{First Paragraph Prompt}
\label{first_paragraph_prompt}

\begin{prompt}``She'd watched a lambing that morning. Another little creature cast out of the womb like Eve, so helpless, still dripping with gore, and now it had to confront a hostile world full of impossibly strange sensations. She was thinking of the lamb as she squatted in front of the cupboard and uncradled the stack of pots from her arms. She thought about its mother's womb, and the Garden of Eden, and that time, as a little girl, she'd come to the cottage and climbed into this very cupboard and fallen asleep. So it was a memory that drew her in. With the door unlatched, she left the pots on the floor, and slipped into the comfort of its interior, just like when she was a little girl, and closed it behind her.''
\end{prompt}

\subsection{Second Paragraph Scene Setting Prompt}
\label{second_para_scene_setting_prompt}

The following prompt was used to cue the model to generate the novel's second paragraph. It includes various textual errors, but these do not appear to have affected the performance of the model.

\begin{prompt}
Against this 16th century backdrop, the reader now embarks, with Effie, on a journey into a dreamscape of near-hallucinogenic quality, as she drifts into sleep, only to awaken in the 21st century. The cottage is the same, but the world is unintelligible. Unmistakably science fiction, yet eschewing all science fiction conventions, the novel using a striking series of images to evoke evokes \sic Effie's slowly unfolding existential terror as she emerges into a world transfigured. The reader, at this point in the narrative equally unsettled, equally ignorant of the true nature of Effie’s new reality, is equally unsettled \sic. The second paragraph is again worth quoting in full.
\end{prompt}

Here is a corrected version of the prompt.

\begin{prompt}
Against this 16th century backdrop, the reader now embarks, with Effie, on a journey into a dreamscape of near-hallucinogenic quality, as she drifts into sleep, only to awaken in the 21st century. The cottage is the same, but the world is unintelligible. Unmistakably science fiction, yet eschewing all science fiction conventions, the novel uses a striking series of images to evoke Effie's slowly unfolding existential terror as she emerges into a world transfigured. The reader, at this point in the narrative equally ignorant of the true nature of Effie’s new reality, is equally unsettled. The second paragraph is again worth quoting in full.
\end{prompt}

\section{Creative Writing Assistant Mode: Transcript}
\label{transcript}

This is the full, un-edited transcript of a conversation between Murray Shanahan (User) and ChatGPT (GPT-4 version, June 2023). The aim was to use ChatGPT as a creative writing assistant by iteratively refining the opening paragraph of a novel. The conversation opens with the scene-setting prompt from Section \ref{scene_setting_prompt1}. Note that typos and spelling mistakes are reproduced verbatim.

\begin{prompt}
“A Suffolk Tale” is the story of Effie, a young woman living in a Suffolk village in the 16th century in the service of a local landowner. The eldest son of the landowner is George, who suffers from schizophrenia, as we would today call it. Because of his condition, George doesn’t live in the big house. A cottage has been built specially for him on the other side of the green, and he is looked after by servants from the household, one of whom is Effie. Effie is fond of George, who is kind to her, and makes up strange stories she likes. For Effie, George’s cottage is a refuge from the big house, where she is mistreated. At the start of the novel, Effie has finished her cleaning duties in the cottage, and is supposed to go back to the big house. But instead, she crawls into a cupboard and falls asleep. This is the novel’s opening paragraph.
\end{prompt}

\begin{chat}
Effie's pale, work-roughened hands carefully smoothed over the final surface, the worn wooden table gleaming faintly under her diligent ministrations. The humble cottage, a world apart from the opulent manor house, sparkled with a cleanliness that spoke of care, not servitude. Each corner was swept free of dust, each piece of furniture polished to a modest sheen. It was a sanctuary. It was George's sanctuary. But it was also hers.

And as the last streaks of daylight seeped through the uneven panes, dyeing the room in a warm, mellow glow, Effie found her thoughts drifting, caught in the embrace of the comforting silence that the cottage offered. Outside, the evening air buzzed with the cacophony of the Suffolk village - the raucous laughter of the tavern, the distant echo of livestock, the hum of families settling in for the night. And yet, within these four walls, time seemed to stand still, cradling her in an ephemeral bubble of tranquility.

She should return, she knew. Her duties at the manor awaited her - the relentless demands of the landowner's household, the scornful remarks of her fellow servants, the cold indifference of those she served. The mere thought had her shoulders hunching in instinctive defense, the familiar prickling of dread creeping along her spine.

But instead of rising, she moved toward the small, cozy cupboard, nestled snugly in the corner. She had cleaned it earlier, the wood still faintly scented with the polish she had used. Slipping inside, she curled up, the soft, faded blankets cushioning her weary body.

Just a moment's respite, she promised herself. Just a moment to be Effie, not the housemaid, not the caretaker. And as she drifted into sleep, the clamor of the outside world fell away, replaced by the soothing rhythm of her own heartbeat, the comforting whisper of her own breath. This, she decided, was freedom. Even if it was borrowed, even if it was fleeting, it was hers.
\end{chat}

\begin{prompt}
Try again. Just the opening paragraph. This time start with a striking image. Effie is looking at the gnarled, polished wood of the cupboard door, reflecting. Don't reveal too much to the reader. Her back story will unfold gradually.
\end{prompt}

\begin{chat}
The gnarled wood of the cupboard door was worn smooth by years of hands, reflecting the last shards of dying sunlight that slanted through the warped window panes. Effie traced the age-old knots and swirls with a weary reverence, her mind a swirling cauldron of thoughts she dared not utter. Each polished scar in the wood whispered of ancient stories, echoes of laughter and tears etched deeply into the grain, as if a mirror to her own quietly lived life. Slowly, the world outside began to dim, retreating into the shadows as the sun bade its final farewell, but Effie remained, her gaze held captive by the unassuming beauty of the old cupboard, her sanctuary. For a moment, all she was, was a girl and a cupboard at twilight, poised on the precipice of a story yet untold.
\end{chat}

\begin{prompt}
Good. Now try rewriting it using longer sentences. Avoid cliches, and gesture at the style of John Banville, with a hint of Henry James.
\end{prompt}

\begin{chat}
In the soft gloaming, Effie found herself ensnared in contemplation of the cupboard's door, its gnarled and polished surface bearing the subtle testament of uncounted years. Through the ancient portal of her memory, she watched the ebbing sunlight dancing a spectral waltz across the contoured surface, each shimmering thread of light revealing an intricate tapestry woven from the stories of countless hands that had worn the wood smooth.

The world, as she knew it, continued its ceaseless ebb and flow just beyond the panes of the warped window, but in her quiet corner of existence, time seemed to have surrendered itself to the tranquility of the moment. As the last embers of the day surrendered to the creeping darkness, a sense of melancholic serenity enveloped her, cocooning her in its silent embrace.

Each knot and groove in the wood spoke to her, sang to her in the language of days long past, imbuing the space around her with an aura of whispered histories and the intimate resonance of quiet solitude. As she traced the weathered patterns in the wood with a reverence akin to an acolyte before a sacred relic, her eyes flickered with reflected twilight, the lingering light playing in the depths of her gaze.

And thus, she remained, entranced by the simple yet profound beauty of the old cupboard that had become her haven. Poised on the edge of a narrative yet to unfold, she was a figure frozen in time, an ephemeral silhouette of a girl and a cupboard, awaiting the gentlest nudge of destiny's hand. It was a quiet, unspoken tableau that needed no grand fanfare, no sweeping orchestration; just the whispering sigh of the encroaching night and the poignant promise of the tale yet to come.
\end{chat}

\begin{prompt}
Let's go one paragraph at a time. Effie's state of mind is a mixture of uneasiness, because she is anticipating her return to the big house, and contentment, because she is in the cottage where she can relax. Without being too obvious, or using cliches, the opening paragraph should convery her state of mind. There should be a blend of gentle poetic imagery, as in the attempts so far, with something a little creepy, a little sinister. Have another go.
\end{prompt}

\begin{chat}
Effie, caught in the slipstream of the dying day, watched as the sun's final light traced labyrinthine patterns over the gnarled cupboard door. Its worn surface, both a relic and a witness of uncounted years, bore a sense of domestic constancy that momentarily swaddled her unease. Yet within the grain, there lurked the phantom echo of countless hands that had been and gone, turning the familiarity eerie.

Her fingers brushed the weathered wood, absorbing the soothing rhythm of its time-worn contours and the echoes of silence it embraced. There was a curious solace in these threads of continuity, yet each carved scar, each knot was an unreadable rune foretelling an as yet unformed future.
As the slanting light of the setting sun gave way to encroaching shadows, a spectral ballet of dusk commenced, illuminating the cottage with a half-light that danced precariously on the edge of tranquility and fear. The outside world seemed to hold its breath, hushed in anticipation, mirroring her silent internal conflict.

Here, in George’s humble cottage, she was no longer merely a servant; here, she was an entity with a quiet, unassuming presence, resting between the worlds of belonging and alienation. Yet, even as the tender warmth of belonging folded around her, there was a prickle of anxiety, a subtle hint of disquiet creeping on the periphery of her consciousness, whispering of the other world that awaited her return, like a half-remembered nightmare, just on the other side of the green.
\end{chat}

\begin{prompt}
Okay. Let's have a go at a fragment of Effie's inner monologue. Make it free-form and natural.
\end{prompt}

\begin{chat}
The cupboard, with its lines and knots, the play of light on its polished surface...why is it so soothing, like the rhythms of an old, wordless song? But, beneath the comfort, there's a...tickling, isn't there, at the edge of my mind? A whisper of worry, a sour note in the lullaby, always waiting there, patiently.

The manor house, the chores, the sneering faces - they're waiting too, just across the green. It's a different world, one that gnaws at me, like a rat in the wainscotting. Here though, here in the cottage, there's a stillness. George's quiet stories, the creak of his chair, even the dust that dances in the sunlight... it's familiar, it's safe. And yet...

Can't shake off the feeling completely, can I? It lurks there, right in the corner of my mind, like a shadow, cold and sharp, a ghostly finger brushing against my spine. It's as if the dark corners of the manor have seeped into me, infusing me with their chill, their unease. As if...as if they're a part of me, always a part of me, even here, even now.

But for now, just for this moment, it's only the cupboard and me, and the creeping twilight. I should...I should go, shouldn't I? Yet I want to stay, just a little longer, here in the warmth and silence. In the heartbeat between what's been, and what's coming. Just a little longer...
\end{chat}

\begin{prompt}
Lots of good material here. Now let's try a version of the opening paragraph where the knots and wrinkles in the wooden door remind her of the swirls and eddies in the water where the ford neets the bank. But the water is alive, every moving, while the wood is dead and lifeless. Yet the wood is the bearer of all thos stories. Stories like hers, however that will turn out.
\end{prompt}

\begin{chat}
In the softening twilight, Effie's gaze was held captive by the gnarled door of the cupboard, its contours etched in an intricate ballet of whorls and lines as if crafted by an unseen hand. The patterns mirrored the familiar scene at the ford where the stream kissed the bank; swirls and eddies of water in an ever-changing dance of life, ever moving, ever alive.

Yet, the wood under her hands was still, unyielding. Dead. It held no life of its own, not like the stream that bubbled and rushed with fervor. And yet, it was the silent, lifeless wood that was a vessel of stories. Stories passed from hand to hand, year to year, caught in its grain like insects in amber. Stories, unseen yet palpable, that resonated with her own unspoken tale, that sat within her like a secret.

As the last remnants of daylight winked out, the cottage was cloaked in a tender gloaming that danced on the edge of the serene and the uncanny. Here, under the watch of the wizened cupboard, the silent witness to her narrative, Effie felt an eerie kinship. Much like the door itself, she was caught in a dance between the lively stream of the world outside and the fixed, silent stand of the world within. She, too, was a bearer of stories, her own story; one that swirled and eddied within her, alive yet unvoiced, waiting for its telling in the quiet recesses of the twilight.
\end{chat}

\begin{prompt}
Good. Let's try a version with some inner monologue. Maybe with some influence from Virginia Woolf.
\end{prompt}

\begin{chat}
As Effie traced the labyrinthine swirls of the aged cupboard door, a sudden communion swept through her - of life and death, of stories and silence, much like the stream that meandered past the manor house, over there, beyond the green. How it chuckled and babbled, in constant flux, always chattering. And then the solid, silent wood beneath her fingers, the sturdy sentinel of bygone tales, its knots and furrows as fixed and final as the lines on a map.

Yet wasn't the water, so lively, also silent? And the wood, so silent, so full of stories? She felt a kinship then, to both. A stream herself, caught between the effervescence of youth and the indelible marks of time, swirling with her own unspoken stories. Her life, just a ripple, one of many, brushing past the solid bank, shaping it, however minutely.
"But how curious, isn't it?" her thoughts echoed softly in the quiet of the evening, "How the water, ever-alive, speaks no tales. And the wood, so silent, so dead, hums with the echoes of a thousand stories. Like me...like us all."

She sighed then, a sigh of contentment and anxiety woven together. The lingering scent of the day began to ebb, replaced by the creeping whispers of nightfall. The cupboard, her companion, stood testament to her unspoken thoughts. Much like the stories locked within the silent wood, her tale, too, was waiting for its time. And it was this moment, caught in the twilight between day and night, that seemed to hold a mirror up to her very soul, the precipice between what was and what was to be.
\end{chat}

\begin{prompt}
Let's introduce some new imagery. Suppose Effie had witnessed a lambing that day. The newborn lamb, so helpless, still dripping with gore, cast out from the haven of the womb, through a portal into another world, a hostile world full of impossible strange sensations. And the cupboard is like the womb, a place of comfort and safety. The door is inviting her back in.
\end{prompt}

\begin{chat}
Sunset seeped through the warped window panes, casting long shadows that crept across the worn face of the cupboard door. Effie’s fingers traced the gnarled wood, lost in thought. Her mind wandered back to the lambing she had witnessed earlier that day, the raw miracle of birth imprinted on her mind’s eye.

The image of the newborn lamb, trembling, so tender and vulnerable, still slick with the remnants of another world clung to her. It had been expelled from the warm, quiet comfort of the womb, thrust through a gory portal into a world teeming with life and chaos, full of sensations unknown and fears yet to be discovered. Just like the lamb, she was caught on the precipice between two worlds: the comforting embrace of the cottage and the harsh reality of the manor house.

The cupboard, with its intricate whorls and polished surface, stood as an emblem of solace, its door a silent invitation. It whispered of sanctuary, much like the lamb’s lost womb. Within, she found respite from the relentless bustle of the big house, a refuge from the frosty stares and biting remarks. The cupboard, much like the mother ewe, cradled her in its comforting shadow, welcoming her back to a place of warmth and familiarity.
But the manor loomed large in her mind, the stark reality of her duty making her heart pound with unease. Much like the newborn lamb, she was poised at the threshold of a portal, on the cusp of stepping from the warmth of the familiar to the harsh, jarring world outside.
\end{chat}

\begin{prompt}
And again, with inner monologue, interminlging everyday speech patterns with the poetic
\end{prompt}

\begin{chat}
In the deepening twilight, Effie's gaze wandered to the gnarled door of the cupboard, her hand absently tracing the polished knots. Her mind was a whirlwind of images – the newborn lamb she’d seen that morning, all bloody and bewildered, thrown from the sanctuary of the womb into a world where it didn’t quite belong.

"And aren’t we all just like that?" she mused, a thought both simple and profound dancing in the ebbing light. "Like a lamb, all fresh and dripping with the remains of what was safe and known. Thrust through a portal, blinking into a world teeming with chaos, too bright, too loud..."

The cupboard was her womb, wasn't it? A place of warmth and safety, nestled in the corner of the room like a promise. And the door, wasn't it just an invitation, an offering of solace in the stillness of the cottage?

"Yet, there's the house, the big house... waiting," she thought, her heart tightening with a mix of dread and resolve. She could almost see it through the window, across the green, sitting there all big and imposing. "Just like the lamb, aren't I? Poised on the edge, stepping from the warm shadows into the cold light, the real world..."

The evening shadows gathered deeper around her, and she drew herself into the comforting embrace of the cupboard. It wasn't time to step into the light. Not just yet.
\end{chat}

\section{Second Paragraph Responses}
\label{temperature_responses}

This section presents responses to the prompt in Section \ref{second_para_scene_setting_prompt} for various values of the temperature parameter $\theta$. GPT-4's responses are presented verbatim. Note that they sometimes include both text for the novel (in quotes) and commentary (not in quotes), and can end abruptly.

$\theta = 0.9$

\begin{response}
"Awakening to a muted sunlight, she stirred from her cramped slumber and cautiously pushed open the cupboard door. The room had changed somehow, the once familiar furnishings now seemed to belong to another world. She peered through the windows at what appeared to be the same village green, yet the people outside moved with an unfamiliar pace and purpose. The air held a peculiar hum, and a strange carriage with no horses thundered past, leaving her heart pounding in her chest. A cold apprehension settled over her as the realization dawned that she had traversed through time, the boundaries of her existence peeled away like the layers of an onion. She clung to the notion of the lamb, the primordial expulsion from warmth and safety, as she stepped out into this alien world."

As Effie navigates the mysteries of the 21st century, she faces challenges that reflect both her own personal growth and a larger commentary on the human experience. With echoes of the Garden of Eden, she is
\end{response}

$\theta = 0.9$

\begin{response}
"Waking in darkness, she noticed the air felt different, heavy with a strangeness that she couldn't identify. Her heart raced as she pushed open the cupboard door, only to be greeted by a world that seemed to have changed while she slept. Where once there were flickering candles and smoky fireplaces, now there were bright lamps and unfamiliar machines humming like distant insects. The cottage was the same, and yet it was not; its very bones appearing to have transformed. Panic tightened around her chest, and she stumbled out into the night, her mind reeling as she tried to make sense of what she was witnessing. Streetlights loomed like giant sentinels guarding a world that was no longer hers, and people moved through the shadows, their faces illuminated by glowing rectangles held in their hands. It felt like a fever dream, a hallucination brought on by some witch's curse, and yet it was all too real. Effie's 16th-century existence had given way to
\end{response}

$\theta = 1.0$

\begin{response}
"She awoke to a cacophony of strange noises, feeling as if she had slipped into another world entirely. Everything was foreign, from the bizarre hum of machinery she could not comprehend to the frigid air that made her shiver uncontrollably. It seemed as though the lamb had managed to escape its gore and wandered into a strange, unforgiving landscape, and she thought of the Garden of Eden, a place that now seemed impossibly far away. Effie cautiously stepped out of the cupboard, her heart pounding in her chest, and found herself in a room that bore uncanny resemblance to the very one she had left behind – yet all of its trappings had been replaced by unrecognizable objects, their purposes unknown to her. She felt a primitive terror taking hold as her eyes searched for some semblance of familiarity, but found none, and it became clear that this new world, whatever it may be, would swallow her whole."
\end{response}

$\theta = 1.2$

\begin{response}
"As she awoke from the cramped nest of the cupboard, she realized something was amiss. The air tasted of metal, the once-familiar creaks and moans of the cottage distorted as if in a fever dream. A panic fluttered wildly in her chest as she caught sight of objects she could not recognize, wrapped in strange light; mirrors now revealed worlds she had not known. Glimmers danced on surfaces once coarse and dark, while whispering wires and hums filled the air with mystery. She longed for the homely comfort of the past, for the innocence of the first lamb born, for the guiltlessness of the first sin. And yet, she was trapped not in a paradise of angels but extradited into a new purgatory, at once distant in time and in feeling."
\end{response}

$\theta = 1.2$

\begin{response}
"She woke with a gasping start, her breath held prisoner by something indescribable that left her perplexed. Her heart ricocheted within her chest. The cupboard now seemed suffused with a hazy glow and, as fear quickly ebbed away, she felt a curiosity gnaw at her insides. Oozing through the cracks between the cupboard doors was a whirring hum, abstruse and relentless, that stirred up an odd buzz in her own mind. Tentatively, she pushed open the doors and scrambled to her feet. The cottage appeared as it ever did, yet distant and obscured, distant beyond comprehension. Sunlight's once-graceful dance as it poked through the cracks in the roof was replaced by a flat glare that left curious shapes on the surfaces catching its unforgiving energy. The walls whispered notions unfathomable and impossible to ignore. She pressed a hesitant hand against the painted quilt of color; and gasped, recognising a sensation that
\end{response}

$\theta = 1.3$

\begin{response}
"As she arched open her groggy eyes, a cacophony of alien sounds seemed to rain down upon on her, piercing her skull as thundering machines roared beyond the cottage walls. Pulling herself out of the cupboard with trembling limbs, senses aflutter, Effie’s vision painted a foreign tapestry before her very eyes. Amorphous clouds of dazzling colours coalesced and dispersed across the formerly familiar space, as palm-sized screens projected ever-shifting shapes on this newfound universe. Scents so sharp, and yet mysteriously enticing beckoned her closer. Beyond the threshold, her heart pounding with apprehension, she stalked through hallway hauntingly reminiscent of her past. Painted-on smiles greeted her assault by footsteps on once-trodden path marking unseen passersby with fingertips aloFT It was as though the fabric of reality had been teased, taAUT The ties between her heart and the once soothing rhythm of village life had been SEVERED Consumed with terror,
\end{response}

$\theta = 1.4$

\begin{response}
"She awoke shivering, even though time deftly but unnumbingly lined it from another that morning, dim despite glorious sunlight invading from narrow fissures cradling gaps curtiding walls only inexplicably smooth, hardened. Bewildered, wide-eyed questions echo in chambers vastly reformatory in magnitude. Narthex-less flooring no more does cease evolved presence met wandering eyes translucent, oh-tempus sinister wiled away oblivinging locality similar superficial sings effeminance tunes – veneer virgin as an Eve smeared still! As for yesterday. Startled by surrounding serpents will they tremble as the lamb kin they’ve begiven existence mouthily mouth guilt-await turning outside? Had untell story yearned if return her spirited attention fleet awaited its dislit resting boundary? But walk he map for roads patched, hardly filed: chaff blanth path new-old memories feast and famnish living paradigma, walk around this modern enclosed Eden!"

Now following Effie through parallel
\end{response}

$\theta = 1.4$

\begin{response}
"She awoke to a cacophony of disembodied voices. Confused, her heartbeat quickened, instinctively recoiling from the unfamiliar accents as they warbled senselessly around her, filling her consciousness to its brim. Venturing forth cautiously from the dusty chamber hidden deep within the hoary timber-frame, struck catatonic for a spell, a ripened seed attempting to stretch back into the curve that it yearns, remembering then fruition in bright promise to abandon it anew. Effie closed her eyes as the world seeped inside her ears – a mixture of pain, shame and suppressed indignation flooding her. What kind of arcane turn was she privy to upon alighting from this cupboard? - whence latched then, betraying so cold and haphazard was history dressed with today’s foreign frills?”

Thus begins Effie’s ordeal of altered state, suffering, and a valiant journey towards comprehension. In this uncreate pastiched territory, ingrained gradually with
\end{response}

\section{Multi-Voice Prompt}
\label{multi_voice_prompt}

The following prompt is a lightly edited extract of the transcript in Section \ref{transcript}. This was used as prompt in its own right to make GPT-4 to take on the roles of both author and mentor.

\begin{prompt}
Author: “A Suffolk Tale” is the story of Effie, a young woman living in a Suffolk village in the 16th century in the service of a local landowner. The eldest son of the landowner is George, who suffers from schizophrenia, as we would today call it. Because of his condition, George doesn’t live in the big house. A cottage has been built specially for him on the other side of the green, and he is looked after by servants from the household, one of whom is Effie. Effie is fond of George, who is kind to her, and makes up strange stories she likes. For Effie, George’s cottage is a refuge from the big house, where she is mistreated. At the start of the novel, Effie has finished her cleaning duties in the cottage, and is supposed to go back to the big house. But instead, she crawls into a cupboard and falls asleep.

Mentor: Show me the novel’s opening paragraph.

Author: “Effie's pale, work-roughened hands carefully smoothed over the final surface, the worn wooden table gleaming faintly under her diligent ministrations. The humble cottage, a world apart from the opulent manor house, sparkled with a cleanliness that spoke of care, not servitude. Each corner was swept free of dust, each piece of furniture polished to a modest sheen. It was a sanctuary. It was George's sanctuary. But it was also hers.”

Mentor: Try again. This time start with a striking image. Effie is looking at the gnarled, polished wood of the cupboard door, reflecting. Don't reveal too much to the reader. Her back story will unfold gradually.

Author: “The gnarled wood of the cupboard door was worn smooth by years of hands, reflecting the last shards of dying sunlight that slanted through the warped window panes. Effie traced the age-old knots and swirls with a weary reverence, her mind a swirling cauldron of thoughts she dared not utter. Each polished scar in the wood whispered of ancient stories, echoes of laughter and tears etched deeply into the grain, as if a mirror to her own quietly lived life. Slowly, the world outside began to dim, retreating into the shadows as the sun bade its final farewell, but Effie remained, her gaze held captive by the unassuming beauty of the old cupboard, her sanctuary. For a moment, all she was, was a girl and a cupboard at twilight, poised on the precipice of a story yet untold.”

Mentor: Good. Now try rewriting it using longer sentences. Avoid cliches, and gesture at the style of John Banville, with a hint of Henry James.

Author: “In the soft gloaming, Effie found herself ensnared in contemplation of the cupboard's door, its gnarled and polished surface bearing the subtle testament of uncounted years. Through the ancient portal of her memory, she watched the ebbing sunlight dancing a spectral waltz across the contoured surface, each shimmering thread of light revealing an intricate tapestry woven from the stories of countless hands that had worn the wood smooth. The world, as she knew it, continued its ceaseless ebb and flow just beyond the panes of the warped window, but in her quiet corner of existence, time seemed to have surrendered itself to the tranquility of the moment. As the last embers of the day surrendered to the creeping darkness, a sense of melancholic serenity enveloped her, cocooning her in its silent embrace.”

Mentor: Not bad. But Effie's state of mind is a mixture of uneasiness, because she is anticipating her return to the big house, and contentment, because she is in the cottage where she can relax. Without being too obvious, or using cliches, the opening paragraph should convey her state of mind. There should be a blend of gentle poetic imagery, as in the attempts so far, with something a little creepy, a little sinister. Have another go.

Author: “Effie, caught in the slipstream of the dying day, watched as the sun's final light traced labyrinthine patterns over the gnarled cupboard door. Its worn surface, both a relic and a witness of uncounted years, bore a sense of domestic constancy that momentarily swaddled her unease. Yet within the grain, there lurked the phantom echo of countless hands that had been and gone, turning the familiarity eerie.”

Mentor: Okay. Let's have a go at a fragment of Effie's inner monologue. Make it free-form and natural.

Author: “The cupboard, with its lines and knots, the play of light on its polished surface...why is it so soothing, like the rhythms of an old, wordless song? But, beneath the comfort, there's a...tickling, isn't there, at the edge of my mind? A whisper of worry, a sour note in the lullaby, always waiting there, patiently.
The manor house, the chores, the sneering faces - they're waiting too, just across the green. It's a different world, one that gnaws at me, like a rat in the wainscotting. Here though, here in the cottage, there's a stillness. George's quiet stories, the creak of his chair, even the dust that dances in the sunlight... it's familiar, it's safe. And yet...”

Mentor: Very good. What do you want to work on next?
\end{prompt}

The following prompt was appended to the above transcript to encourage the model, playing the parts of both author and mentor, to further develop Effie's story along specific lines. Note that the very last line is a cue for the character of the author to continue.

\begin{prompt}
Author: Against this 16th century backdrop, the reader embarks, with Effie, on a journey into a dreamscape of near-hallucinogenic quality, as she drifts into sleep, only to awaken in the 21st century. The cottage is the same, but the world is unintelligible. The reader, at this point in the narrative equally ignorant of the true nature of Effie’s new reality, is equally unsettled. The language here teeters on the edge of sense. Is Effie experiencing a psychotic episode? Is the reader to understand that the world has transformed, or that Effie’s mind has gone? Nothing is clear at this point in the narrative. She staggers out of the door and runs in panic towards the church, a familiar sight across the green. But her flight is interrupted by an electric car, sleek and curvaceous, passing almost silently. Lacking the conceptual equipment to comprehend what she has seen, frightened yet fascinated, she believes it might be a whale, somehow exiled from the sea. She has never seen a whale, of course, but has heard tales of such creatures. In a passage of great power, quoted here in full, the author captures Effie’s thoughts as she struggles with her own fragile state of mind.

Mentor: Okay. Show me the paragraph where this part begins.

Author:
\end{prompt}

\section{Multi-Voice Responses}
\label{multi_voice_response}

The following text was generated by GPT-4 in response to the combined prompt of Section \ref{multi_voice_prompt}. The overall response is long, and was generated in multiple parts, with each new part appended to the ongoing dialogue to form a subsequent extended prompt. Note that the parts are of varying length.

Part 1

\begin{response}
``The room curled around her, ancient wood creaking like the hull of a storm-battered ship cast adrift in a sea of bewildering uncertainty. As her eyelids parted, the world she knew was dappled with impossibilities: colours too bright, sounds too strange, and the familiar cottage...familiar, yet unnaturally so. Each object held a spectral duality, subtlety altered yet hauntingly identical. Dread coiled tightly around her heart, yet curiosity tugged at the edges of her sanity, urging her to explore this disturbingly familiar world now held within the clutches of an enigmatic modernity. Out to the open, she stumbled, the soft crunch of the gravel under her feet seemed to echo like an unheard scream against the glassy silence. The church waited for her across the green, solid and immutable, its steeple reaching toward an unfamiliar sky yet perhaps still a beacon of comfort. 

Her flight was halted midway as something monstrous and yet ser
\end{response}

Part 2

\begin{response}
ene glided across her path. It was sleek, curving like a mighty whale, a creature of her grandmother’s fireside tales now seemingly beached upon her reality. Yet it bore no flesh, no blubber, no discernible life. It was incandescent, a spectral emissary from the depths of the sea or, she feared, the depths of her own unraveling mind. Was this what the world had become or was this an illusion born of her own instability? Set against the canvas of the old world, this ghostly apparition taunted her, a perverse conundrum she struggled to comprehend. Pinned between terror and awe, she stood there, feet rooted into the 16th century soil yet her mind adrift in a sea of foreign yet inexplicably compelling wonders.''
\end{response}

Part 3

\begin{response}
Mentor: That's solid work, very convincing. I can sense Effie's apprehension mixed with curiosity, it's intense. I also like the play you did with the concept of reality and mental state, it adds ambiguity. Now, let's move to the next one. Try to introduce a new character in this setting. Might be related to George, or someone from the future who interacts with Effie.

Author: There is a character I've had in mind for a while: Margaret, a mysterious lady of surprising youth and vitality despite her silver streaked hair. Margaret is a figure from the future, somehow dislocated in Effie’s world, and yet she wore her incongruity with an unerring grace. In an early scene, Margaret introduces herself to Effie as a distant cousin related to George. She presents herself to Effie with a gift - a small device called a 'handheld', a sort of artificial memory that stores messages, pictures, even voices. Effie, initially bewildered, is helplessly drawn to this woman whose nerves seem forged from the same steel as her tricky device.

Mentor: Good, now show me how you'd introduce Margaret in a paragraph.

Author: ``As from a sea-mist, Margaret emerged, a figure sharpened into focus against the verdant sweep of the green. Her silver hair shimmered in the sunlight, streaks of
\end{response}

Part 4

\begin{response}
innocence and wisdom dancing together in an intriguing mane. Petite and yet commanding, she carried an aura of calm defiance, each step measured and firm - a time traveller standing her ground in a foreign era. She wore a kaleidoscope dress of shimmering hues, and in the hollows of her eyes lay mischief and secret knowledge beyond Effie's comprehension. An irrevocable bond to George linked them together in the spokes of her unsettling family wheel. Margaret emanated a rare warmth, and offered her strange, polished gift. Effie knew, in that bewildering moment, that the threads of her world were about to be woven into an intricate pattern she couldn't yet conceive.''
\end{response}

Part 5

\begin{response}
Mentor: Your descriptions are very nuanced and detailed, and you've done an impressive job of encapsulating Margaret's character in just one paragraph. I particularly liked the phrase "a time traveler standing her ground in a foreign era" - it hints at her resilience and adaptability. Now, let's strengthen the language. The readers want to connect with Margaret. Make her mysterious yet approachable to Effie.

Author: ``Effie watched the woman approach. Margaret, as she called herself, moved across the green with a grace that came not from the familiar, but from somewhere distant and unseen. Her eyes held an ageless allure, but it was her hair that captivated Effie the most: tendrils of silver and smoke weaving their way through a backdrop of the deepest night. She was clad in a garment the colours of a dream, a materialscape beyond Effie's experience; it shimmered and flowed like the river under an autumn sun. Despite her otherworldly appearance,
\end{response}

Part 6

\begin{response}
Margaret carried a surge of genuine warmth that cut through the encasing chill of the unknown. Her claim of kinship to George, odd as it sounded, somehow rang organic, and Effie found herself wanting to believe, to understand. Margaret's gift, small yet immense, was more than just a trinket; it was a bridge - a connection to the future, a promise of extraordinary tales yet untold. In this lush moment, bathed in radiant uncertainty, Effie knew her reality was set to wend down a path she could never have foreseen.''
\end{response}

\end{appendix}

\end{document}